\documentclass[nohyperref]{article}

\usepackage{microtype}
\usepackage{graphicx}
\usepackage{subcaption}
\usepackage{booktabs} % for professional tables

\usepackage{hyperref}

\usepackage[accepted]{icml2022}

\usepackage{amsmath}
\usepackage{amssymb}
\usepackage{mathtools}
\usepackage{amsthm}
\usepackage[version=4]{mhchem}

\usepackage[capitalize,noabbrev]{cleveref}

\theoremstyle{plain}

\theoremstyle{definition}

\theoremstyle{remark}

\usepackage[textsize=tiny]{todonotes}

\icmltitlerunning{Towards Massive Public Datasets in Scientific ML}

\begin{document}

\twocolumn[
\icmltitle{The Bearable Lightness of Big Data: \\ Towards Massive Public Datasets in Scientific Machine Learning}



\icmlsetsymbol{equal}{*}

\begin{icmlauthorlist}
\icmlauthor{Wai Tong Chung}{su}
\icmlauthor{Ki Sung Jung}{sandia}
\icmlauthor{Jacqueline H. Chen}{sandia}
\icmlauthor{Matthias Ihme}{su,slac}

\end{icmlauthorlist}

\icmlaffiliation{su}{Department of Mechanical Engineering, Stanford University, Stanford, CA 94305, USA}
\icmlaffiliation{sandia}{Combustion Research Facility, Sandia National Laboratories, Livermore, CA 94550, USA}
\icmlaffiliation{slac}{SLAC National Accelerator Laboratory, Menlo Park, CA 94025, USA}

\icmlcorrespondingauthor{Wai Tong Chung}{wtchung@stanford.edu}

\icmlkeywords{Machine Learning, ICML}

\vskip 0.3in
]



\printAffiliationsAndNotice{} % otherwise use the standard text.

\begin{abstract}
In general, large datasets enable deep learning models to perform with good accuracy and generalizability. However, massive high-fidelity simulation datasets (from molecular chemistry, astrophysics, computational fluid dynamics (CFD), \emph{etc.}) can be challenging to curate due to dimensionality and storage constraints. Lossy compression algorithms can help mitigate limitations from storage, as long as the overall data fidelity is preserved. To illustrate this point, we demonstrate that deep learning models, trained and tested on data from a petascale CFD simulation, are robust to errors introduced during lossy compression in a semantic segmentation problem. Our results demonstrate that lossy compression algorithms offer a realistic pathway for exposing high-fidelity scientific data to open-source data repositories for building community datasets. In this paper, we outline, construct, and evaluate the requirements for establishing a big data framework, demonstrated at \url{https://blastnet.github.io/}, for scientific machine learning.

\end{abstract}

\section{Introduction}

\begin{figure*}[htb]
\vskip 0.2in
\begin{center}
\centerline{\includegraphics[width=\textwidth]{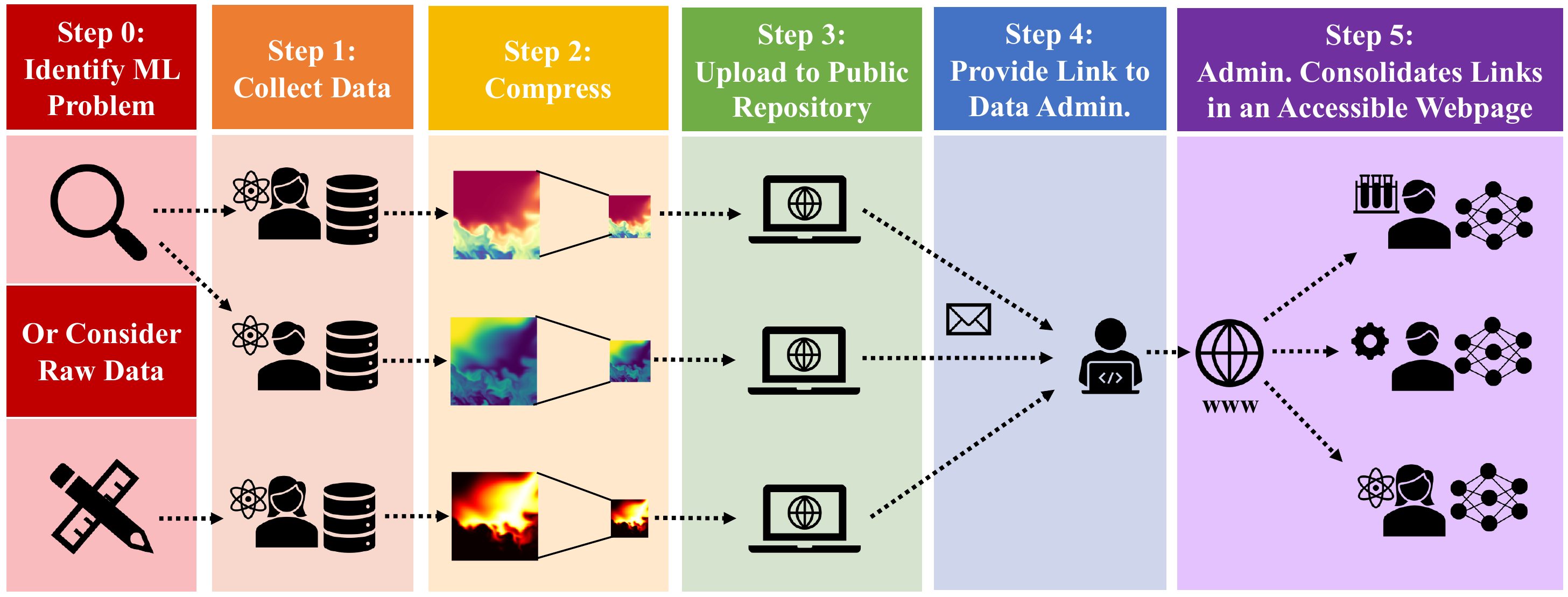}}
\caption{BLASTNet: A path towards public datasets in scientific ML. URL: \url{https://blastnet.github.io/}.  }
\label{fig:proposal}
\end{center}
\vskip -0.2in
\end{figure*}
Accuracy and generalizability are the requirements of predictive machine learning (ML) models. One way to achieve this is to rely on a wealth of sufficiently high quality data~\cite{Sun2017RevisitingUE}. In fields such as computer vision, massive and diverse datasets~($\sim$170 GB, 1.4M images, 1,000 classes) such as  ImageNet~\cite{imagenet_cvpr09}, which is shared via Kaggle~\cite{Kaggle}, have enabled  deep learning models~\cite{resnet_cvpr16} to outperform  human capabilities in image recognition~\cite{russakovsky2015imagenet}. 

In contrast, high-fidelity simulation datasets found in the natural and applied sciences, such as the Johns Hopkins Turbulence Database~\cite{li2008public} are not as diverse (9 simulation cases), but are orders of magnitude greater in size ($\sim$500 TB) due to increased dimensionality and grid resolution requirements.
As such, open-source public ML data repositories such as Kaggle (with a size limit of $\mathcal{O}(100)$~GB per dataset) are not feasible. Instead, significant resources and infrastructure must be dedicated towards building and maintaining data storage facilities. 
Since access to scientific data can be limited, many fields, including material sciences~\cite{Zhang2018},  experimental chemistry~\cite{Thawani_2020}, and the aforementioned flow physics,  have applied ML in the \textit{small data} regime, where ideas such as  knowledge-guided ML~\cite{Karniadakis2021} are popular.

In fields where high-fidelity simulations are prevalent, the wealth of data required to operate in the \textit{big data} regime does exist. For example, \citet{IHME2022101010} identified over 200 high-fidelity simulation cases that could serve as a foundation for a big dataset for turbulent reacting flows.  Thus, if these challenges in data storage can be overcome, ML in the natural and applied sciences could more effectively leverage advances from the broader ML community -- which focuses on big data, big models~\cite{YuanBigModel2022}, and foundation models~\cite{BommasaniFoundation2021} -- towards predictive tasks in scientific problems. We must note that both small and big data paradigms do not necessarily compete, and that effective data-driven models, in fields such as material science~\cite{LuPIMLResnet2017}, flow physics~\cite{wu2018physics}, and astro-physics~\cite{KHAN2020135628}, have been developed by combining ideas from both paradigms.

Here, we propose lossy compression methods~\cite{Capello19Lossy} towards compressing data into tractable sizes, suitable for sharing via public repositories, at the cost of introducing errors (typically controllable via error-bounded algorithms) to a dataset. This is complemented by a recent study~\cite{Northcutt2021} which demonstrated that ImageNet and other popular benchmark datasets contain up to 10\% label errors. Despite these errors in training data, ML continues to perform with remarkable accuracy because modern deep learning algorithms are inherently robust to noisy data~\cite{Rolnick2017DeepLI,Mahajan2018}. This means that lossy compression could be applied towards mitigating storage limits in public repositories. 

In this work, we propose a realistic pathway for building high-fidelity scientific datasets required to operate in the big data regime. This Bearable Large Accessible Scientific Training Network-of-Datasets (BLASTNet) framework  -- combining lossy compression, community outreach, and public repositories -- is summarized in \Cref{fig:proposal}. A preliminary step before data collection involves either (i) identifying target supervised learning problems, where labels can be defined, or (ii) choosing to simply share raw scientific data. Next, the data is collected and compressed into a consistent data format at a desired level of error. Then, the compressed dataset ($\mathcal{O}(100)$ GB each) from different scientific investigators  can be uploaded onto a  public ML repository. A link and description of the dataset can then be shared to a data administrator, who curates the links and metadata from the network of distributed datasets on a community-hosted webpage. In this work, we present a proof-of-concept of this webpage~\cite{BLASTWeb} on \url{https://blastnet.github.io/}.

This webpage also provides tutorials for sharing/accessing scientific data and provides standards for the shared data. A  discussion forum is hosted to receive community feedback and to provide user support. To ensure that fair attribution is provided in this open-source project, a version update will be applied each time a new dataset is contributed so that  each individual contributor is included into BLASTNet's author list, which is a common practice in open-source software~\cite{Goodwin2018Cantera:Processes}. For the first iteration of BLASTNet, we envision a network-of-datasets for high-fidelity simulation data of reacting and non-reacting flow configurations, covering $\sim$100 different configurations with a total of $\sim$1000 different snapshots in order to curate sufficiently massive and diverse datasets, with later versions considering other forms of scientific data.

This big scientific data framework relies on the (i) the size and quality of the compressed data, and (ii) the robustness of deep learning models to errors introduced during lossy compression. To address these concerns, we quantify the errors and reduction offered by a lossy compression algorithm, SZ2~\cite{LiangSZ22018}, and  demonstrate that deep learning is still effective after training on lossy data extracted from a petascale turbulent reacting flow simulation, in a semantic segmentation problem. Within turbulent reacting flows, this type of classification can be useful for detecting catastrophic rare events~\cite{CELLIER2021111558}, optimizing numerical computations~\cite{chung2020dataassisted}, and identifying combustion regimes~\cite{WAN2020268}. In other scientific fields, semantic segmentation have been explored for processing data extracted from microscopes~\cite{Ronneberger2015}, radio-astronomical measurements~\cite{Pino2021}, and neutrino experiments~\cite{neutrino2021}.
The present data is further described in \Cref{sec:data}, while the lossy compression algorithm and the 3-D convolutional neural network (CNN) employed are detailed in \Cref{sec:method}. 
We present our results and conclusions in \Cref{sec:results} and  \Cref{sec:conclusion}, respectively.

\begin{figure*}[htb!]
% \vskip 0.2in
\begin{center}
\centerline{\includegraphics[width=\textwidth]{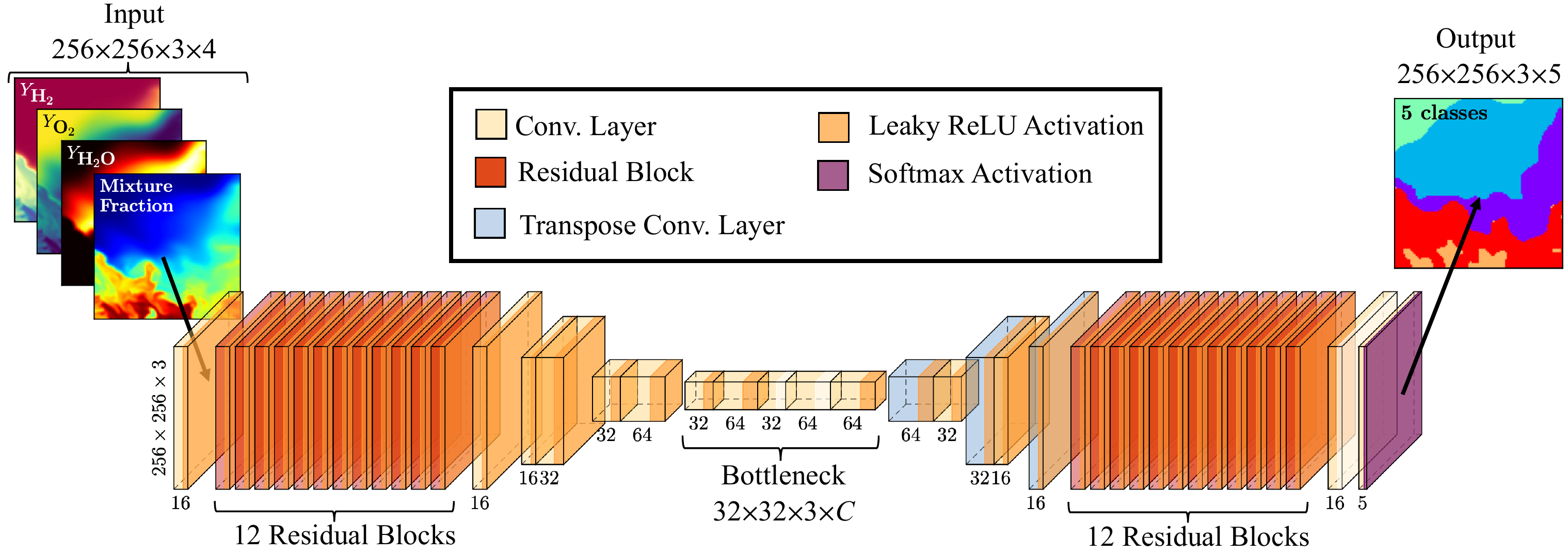}}
\caption{Present 3-D CNN architecture. Number of filters per layer $C$ are described at the bottom of each layer.}
\label{fig:cnn}
\end{center}
\vskip -0.2in
\end{figure*}

\section{Data Description}
\label{sec:data}
A three-dimensional direct numerical simulation (DNS) dataset for a turbulent lifted hydrogen jet flame in heated co-flow air \cite{JUNG2021111584} is used in this study. This simulation data was generated by solving the compressible Navier-Stokes equations, along with species continuity and total energy conservation equations, using a high-order numerical solver~\cite{Chen_2009}, with a detailed  9-species and 21-reaction  hydrogen-air chemical mechanism~\cite{Li04}.

This petascale simulation of reacting flow in a slot-burner configuration consists of 1.28 billion grid points ($2000\times1600\times400$) with 12 conserved quantities for the flow-fields. A single snapshot of this data is slightly more than 100 GB in size. A diluted hydrogen fuel is issued from the central jet with a jet Reynolds number of 8,000. The central jet is surrounded on either side by co-flowing heated air streams with an inlet temperature of 850 K. The computational domain size is 30 $\times$ 40 $\times$ 6 mm$^3$ in the streamwise, $x$-, transverse, $y$-, and spanwise, $z$- directions, respectively. A uniform grid size of 15 $\mu$m is placed in $x$- and $z$- directions while and algebraically-stretched grid is adopted in the $y$- directions.

In the present study, a sub-region (with 60M cells) of the DNS field (i.e., a left half branch of the lifted jet flame) is sampled to evaluate the lossy compression algorithm. From this data, we extract four flow features, and generate five classes of labels from this dataset, and subdivide the data into 268 sub-volumes, each with $256\times256\times3$ cells and four channels for the features in the input flow-field. Note that since this configuration is homogeneous in the spanwise direction, 3 cells in the $z$-axis is sufficient for preserving spatial information in these subvolumes.  The features consist of mass fractions of major chemical species and   mixture fraction\footnote{The mixture fraction can be thought of as reduced dimension of chemical composition of a reacting fuel-air mixture, with values of 0 and 1 corresponding to the amount of material originating from the oxidizer and fuel streams, respectively.} as defined by ~\citet{BILGER197687},
i.e. \{$Y_\text{\ce{H2}}$, $Y_\text{\ce{O2}}$, $Y_\text{\ce{H2O}}$, $Z$\}, and are normalized with a min-max scaler prior to training. The classes consist of premixed flame, non-premixed flame, pure fuel, pure air, and unburned fuel-air mixture which can be generated directly from the magnitudes and gradients~\cite{YAMASHITA199627} of the input features. Train, validation, and test sets are split in a typical 60:20:20 fashion, with random rotation and random flipping used to further augment the train set.

\section{Methods}

\label{sec:method}
\subsection{Deep Learning Model}
The architecture of the 3-D convolutional neural network (CNN) employed is also shown in \Cref{fig:cnn}. This architecture is based on the work of \citet{Glaws2020AE}, and is known to perform effectively in flow physics problems, with the input size of the present model modified to $256\times256\times3\times4$, the minimum number of filters per hidden layer increased to 16, and the filter width reduced to 3. 12 residual blocks are placed before and after an autoencoder network, with a softmax output activation for 5 classes used together with a categorical cross-entropy loss function to solve the present semantic segmentation problem. This network contains 93 layers and approximately 1M trainable parameters, with weights initialized via Xavier initialization~\cite{glorot2010understanding}. Train and validation  procedures are shared in \Cref{app:trainval}.

\subsection{Lossy Compression Algorithm}
We employ the SZ2 compressor~\cite{LiangSZ22018}, which combines  curve-fitting, the Lorenzo predictor, and data quantization, tailored for compressing a wide range of scientific data including measurements from seismic imaging and X-ray, as well as simulation data in molecular dynamics, cosmology, and flow physics.
In principle, this compression algorithm (i) partitions field variables into neighborhoods, (ii) iteratively searches for approximate regression functions that can describe each neighborhood with a guaranteed error-bound, and (iii) stores the quantized regression coefficients of the function and indices of the field variables for reconstructing the data during decompression. Since the quantized coefficients and indices are much smaller than the original field variables, the data can be compressed more effectively than lossless compression algorithms.

For this study, we consider the point-wise relative error method~\cite{LiangPWSZ2018} within this compressor, which  guarantees that the lossy error in each cell does not exceed a  user-defined percentage of the compressed value. This method results in a well-defined measure of quality for describing any shared scientific datasets in a public repository, and is especially useful for maintaining the fidelity of lossy compressed data with large variances, such as with flow velocity and  mass fractions of minor chemical species such as OH.

\section{Results}
\label{sec:results}
\subsection{Effects of Lossy Compression on Data}
\begin{figure}[htb!]
% \vskip 0.2in
\begin{center}
\centerline{\includegraphics[width=\columnwidth]{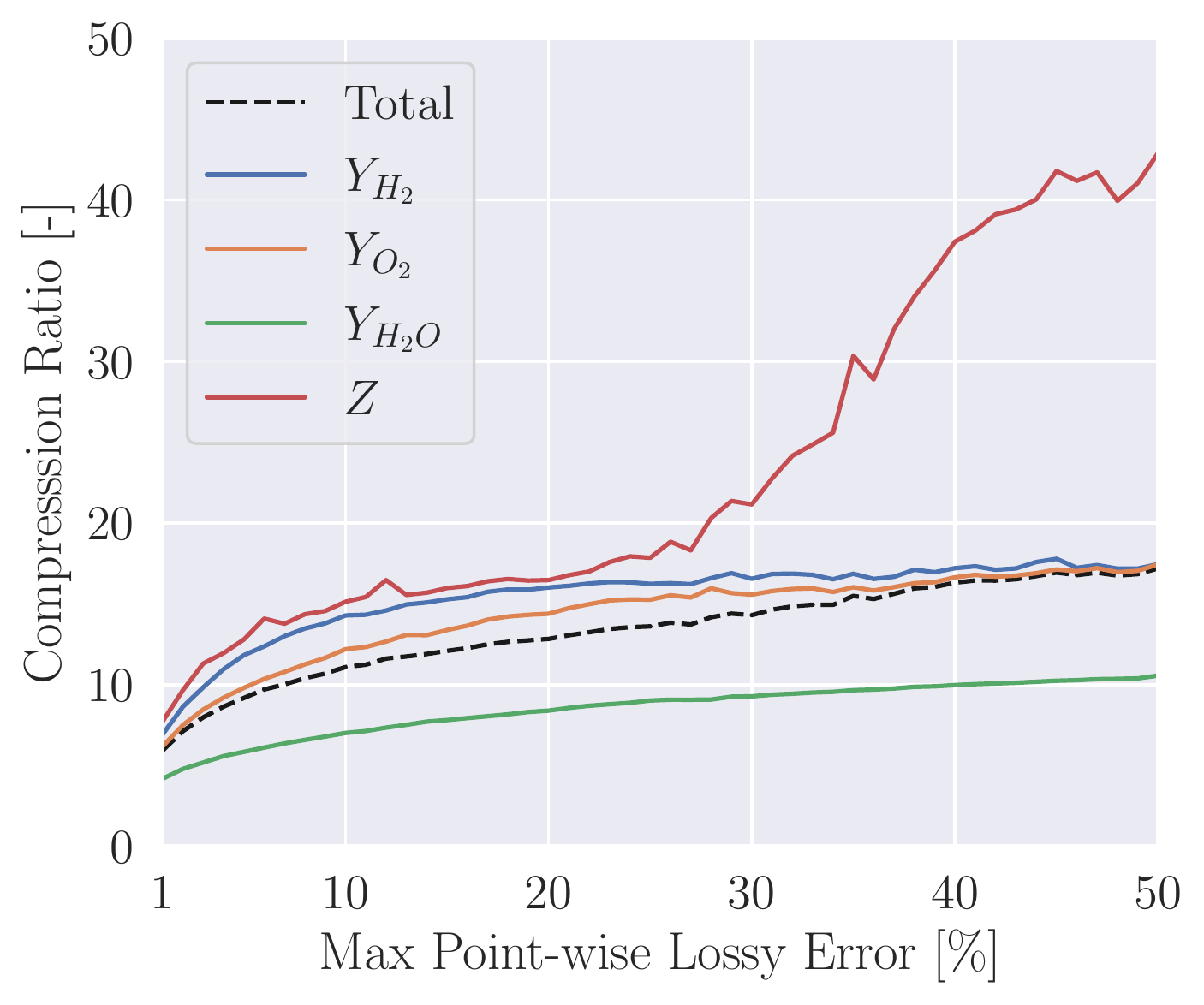}}
\caption{SZ2~\cite{LiangSZ22018} compression ratio of features of the entire learning set at different specified maximum point-wise error.}
\label{fig:sz_cr}
\end{center}
\vskip -0.2in
\end{figure}

We first compress 3-D scalar fields from the entire learning set with SZ2. \Cref{fig:sz_cr} demonstrates the total compression ratio\footnote{Compression ratio is defined as the original file size divided by the compressed file size.} from 1\% to 50\% max point-wise error ranges from 6- to 16-fold compression. Even if we consider only the smallest compression ratio seen in compressing $Y_\text{\ce{H2O}}$, a 5-fold compression of the raw $\sim$100 GB petascale simulation dataset, would enable at least 4 snapshots of this data to be shared as a single dataset on Kaggle. Data compression could be repeated on other flow configurations, and shared via the framework presented in \Cref{fig:proposal} for building a distributed ML training dataset.   

\begin{figure}[htb!]
% \vskip 0.2in
\begin{center}
\centerline{\includegraphics[width=\columnwidth]{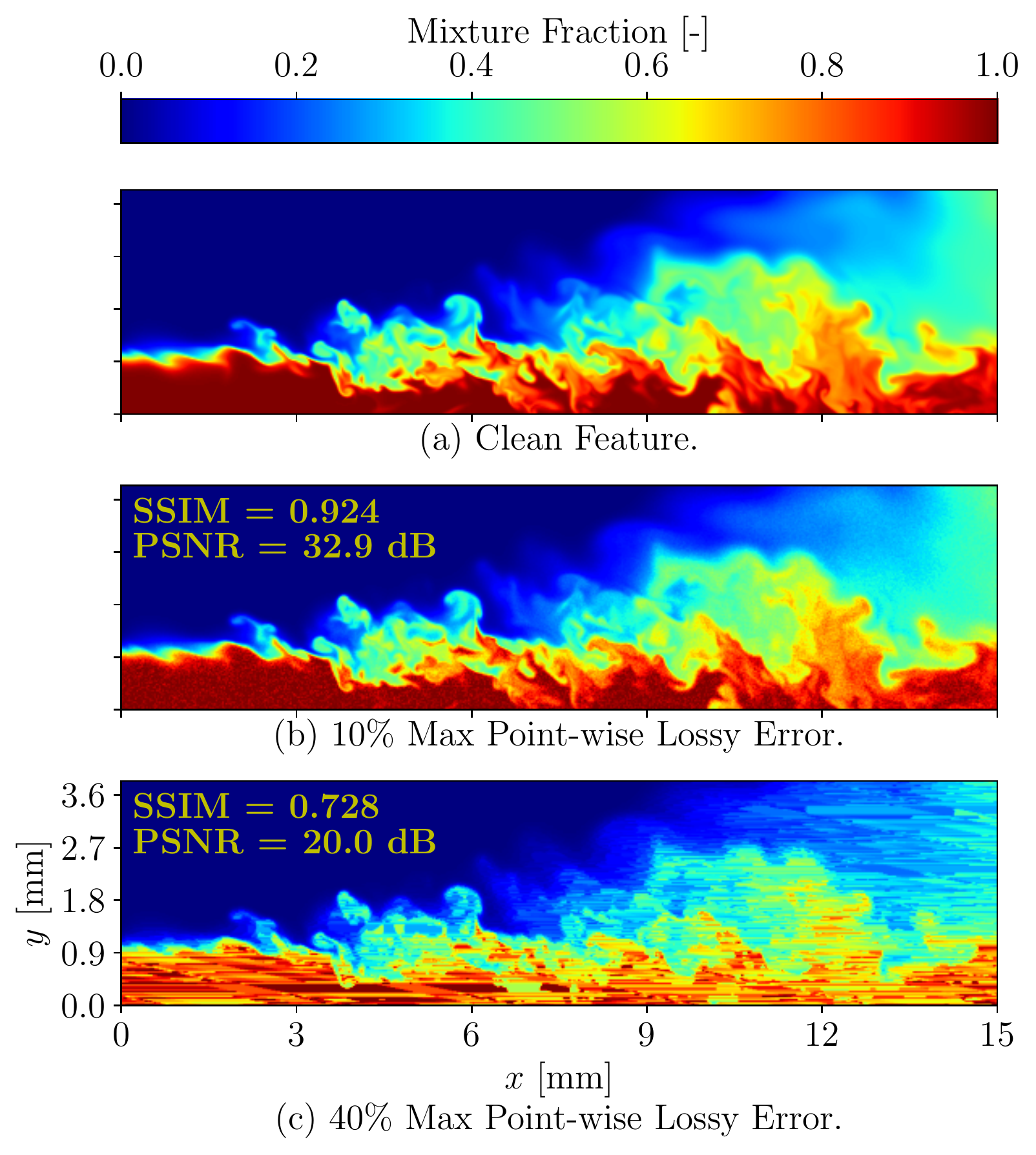}}
\caption{ A feature (mixture fraction) from the train set at different levels of maximum point-wise error specified during compression. Quality metrics such as peak-signal-to-noise-ratio (PSNR) and structural similarity index measure (SSIM) are included.}
\label{fig:feature_lossy}
\end{center}
\vskip -0.2in
\end{figure}

Next, we decompress the compressed data and evaluate errors introduced by the lossy compressor. \Cref{fig:feature_lossy} compares mixture fraction $Z$ at different levels of maximum point-wise lossy error, with the original clean feature. Image quality metrics \cite{Hore2010} such as peak-signal-to-noise-ratio\footnote{Higher PSNR means higher image quality. Note that the highest possible value for PSNR is 48 dB for 8-bit integers, and  760 dB for single-precision floating points.} (PSNR) and structural similarity index measure\footnote{Higher SSIM means higher image quality. SSIM is bounded between -1 and +1. } (SSIM) are shown to decrease with increasing compression, with large field distortions observed with 40\% max point-wise error (\cref{fig:feature_lossy}c).

\begin{figure}[htb!]
% \vskip -0.2in
\begin{center}
\centerline{\includegraphics[width=\columnwidth]{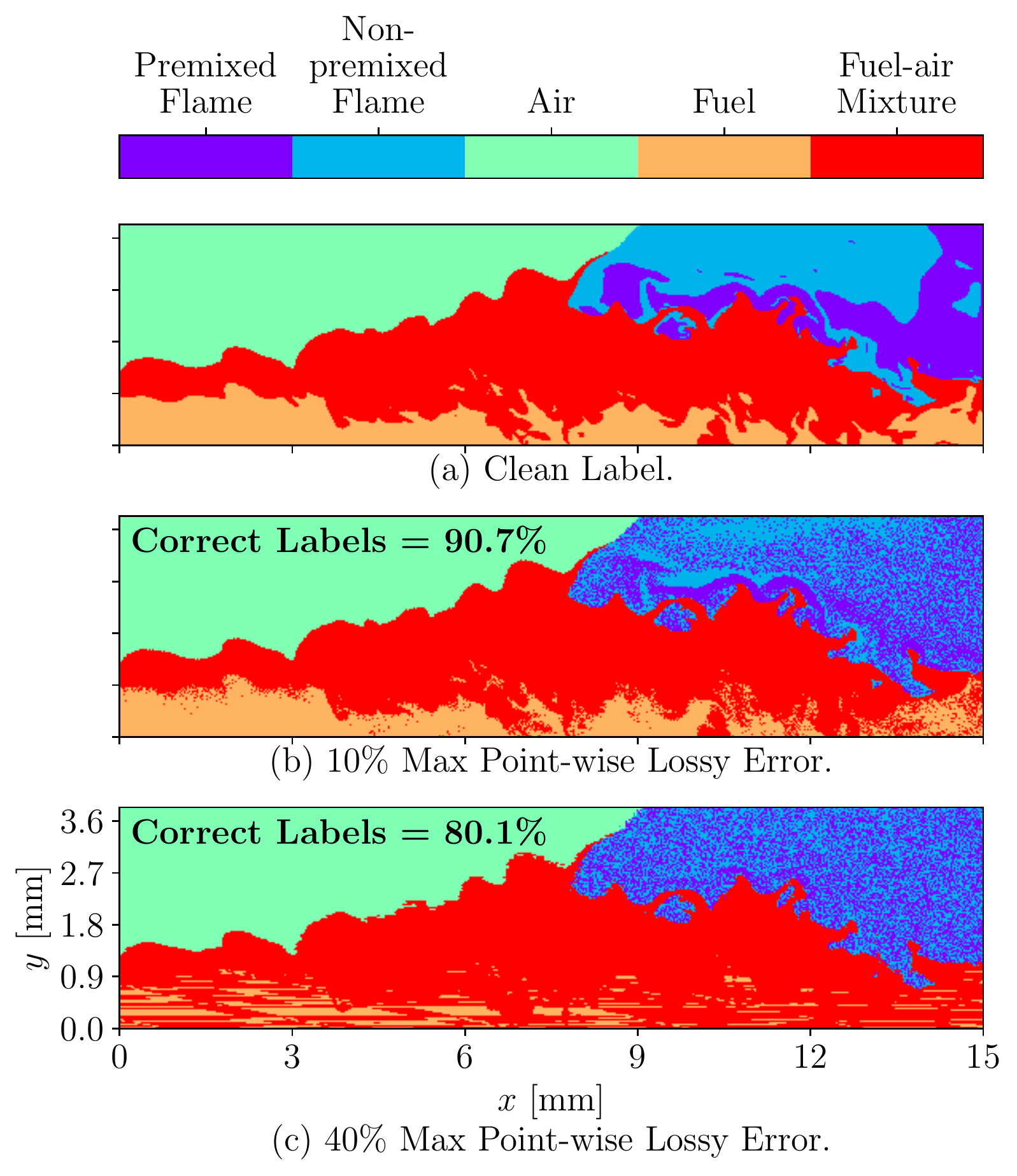}}
\caption{Training labels generated from lossy data at different levels of maximum point-wise error specified during compression.}
\label{fig:label_lossy}
\end{center}
\vskip -0.2in
\end{figure}

After decompressing the scalar fields, we generate the training labels for the different lossy data, as described in \Cref{sec:data}. \Cref{fig:label_lossy}  compares the five classes at different levels of maximum point-wise error, with the original clean training label. At  \Cref{fig:label_lossy}b, significant noise is seen especially in the premixed and non-premixed flame regions at 10\% max point-wise lossy error, with a 9.3\% total label error introduced to the data. This noise is present because scalar gradients, used in generating the flame labels~\cite{YAMASHITA199627}, are not necessarily preserved adequately after lossy compression. \Cref{fig:label_lossy}b shows that the fuel labels become especially distorted at 40\% max point-wise lossy error, with a total label error 19.9\%. 

\subsection{Train with Lossy data, and Test on Clean Data}
In general, validation and test sets do not come from the same distribution as the training data, and are usually sampled to represent data encountered after deployment. Thus, in the big data framework proposed in  \Cref{fig:proposal}, we envision a scenario where large quantities of lossy compressed training data can be easily obtained from public repositories, with small quantities of clean test and validation data sampled personally by a user. 

In this study, we explore effects of training with lossy data, and testing and validating on clean data. This task can be further subdivided into two scenarios: (i) where lossy features are shared into the repository with clean labels, and (ii) lossy labels are generated from lossy data obtained from the repository (such as with \Cref{fig:label_lossy}). The former scenario is encountered where a specific target supervised learning problem (such as with ImageNet for image recognition) has been identified. In this case, lossy compression does not necessarily need to be employed to the labels, since the dimensionality of labels are much smaller than features. This could be more beneficial than extracting potentially noisy labels from lossy scientific data in the latter scenario, especially since ML methods are well-known to be more robust to feature noise than label noise~\cite{Zhu2004}. 
\subsubsection{Lossy Features and Clean Labels}

\begin{figure}[htb!]
        \centering
            \begin{subfigure}{0.95\columnwidth}
        \centering            \includegraphics[width=\columnwidth]{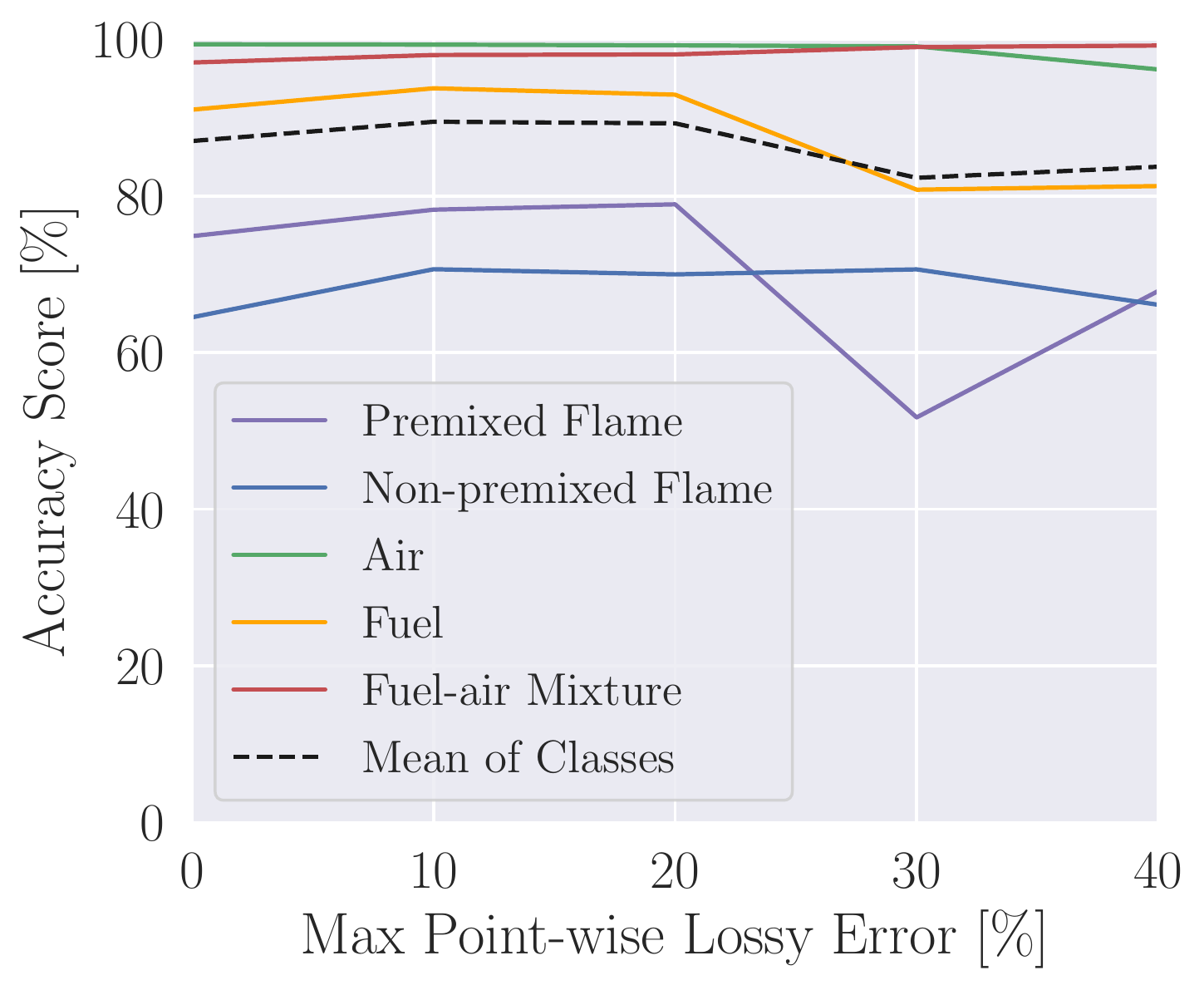}
        \caption{Model trained on lossy features and clean labels, and tested on clean and uncompressed features.}
\label{fig:clean_acc}
    \end{subfigure}
    \begin{subfigure}{0.95\columnwidth}
        \centering            \centerline{\includegraphics[width=\columnwidth]{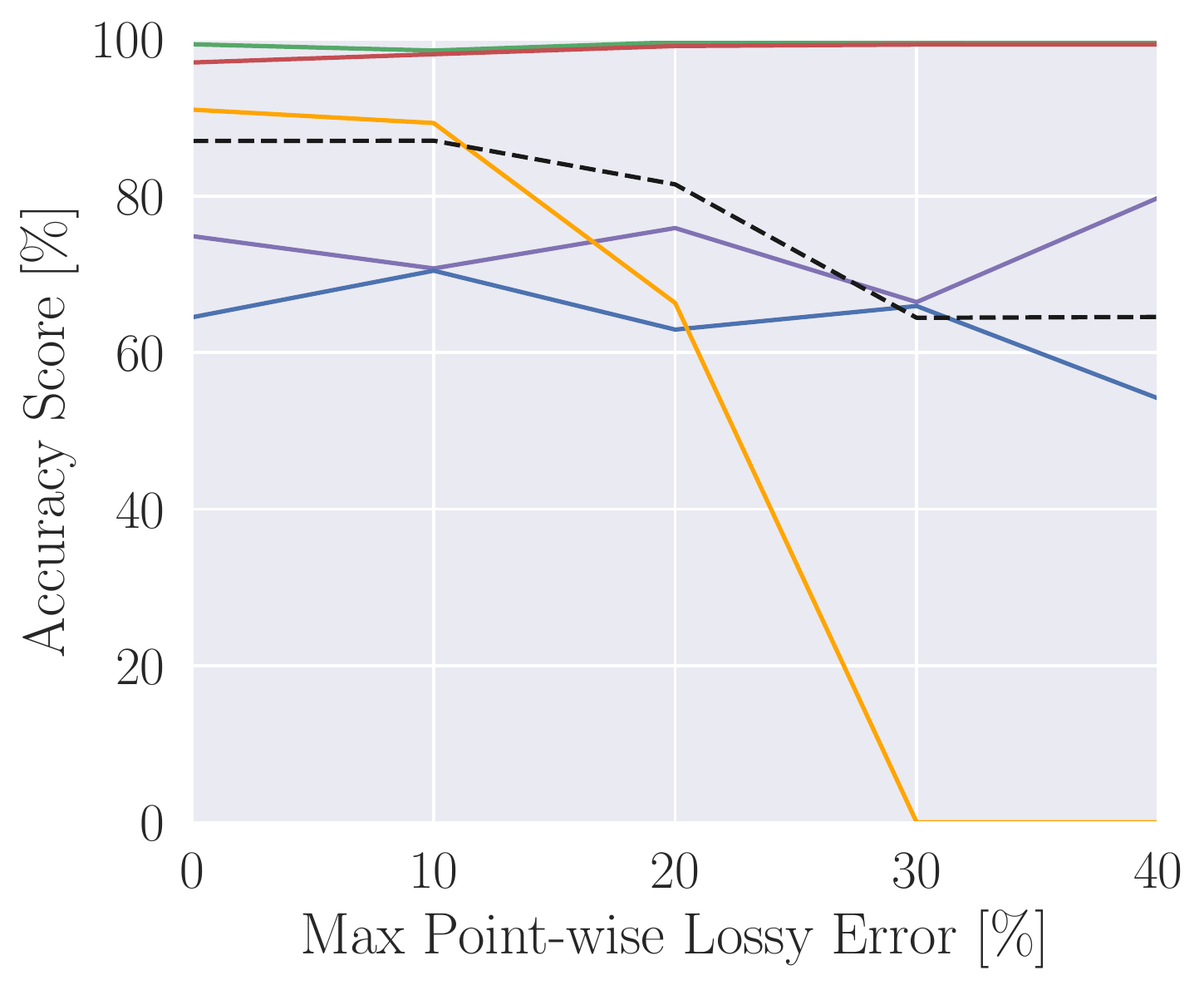}}
\caption{Model trained on lossy features and lossy labels, and tested on clean and uncompressed features.}
\label{fig:noisy_acc}
    \end{subfigure}
    \caption{Class accuracy score at different levels of maximum point-wise error specified during compression. }
    \vskip -0.2in
\end{figure}

\Cref{fig:clean_acc} compares class-specific accuracy scores, along with the mean of these scores, for different levels of maximum point-wise lossy errors, when training with lossy features and clean labels. Mean accuracy score of 87\% is seen in the baseline case of 0\% lossy error, which is typical in a semantic segmentation problem~\cite{Ronneberger2015}. The mean accuracy scores are seen to be robust up to 20\% max point-wise lossy error, which corresponds to a 13-fold compression in the original data.

\begin{figure*}[htb!]
% \vskip 0.2in
\begin{center}
\centerline{\includegraphics[width=0.9\textwidth]{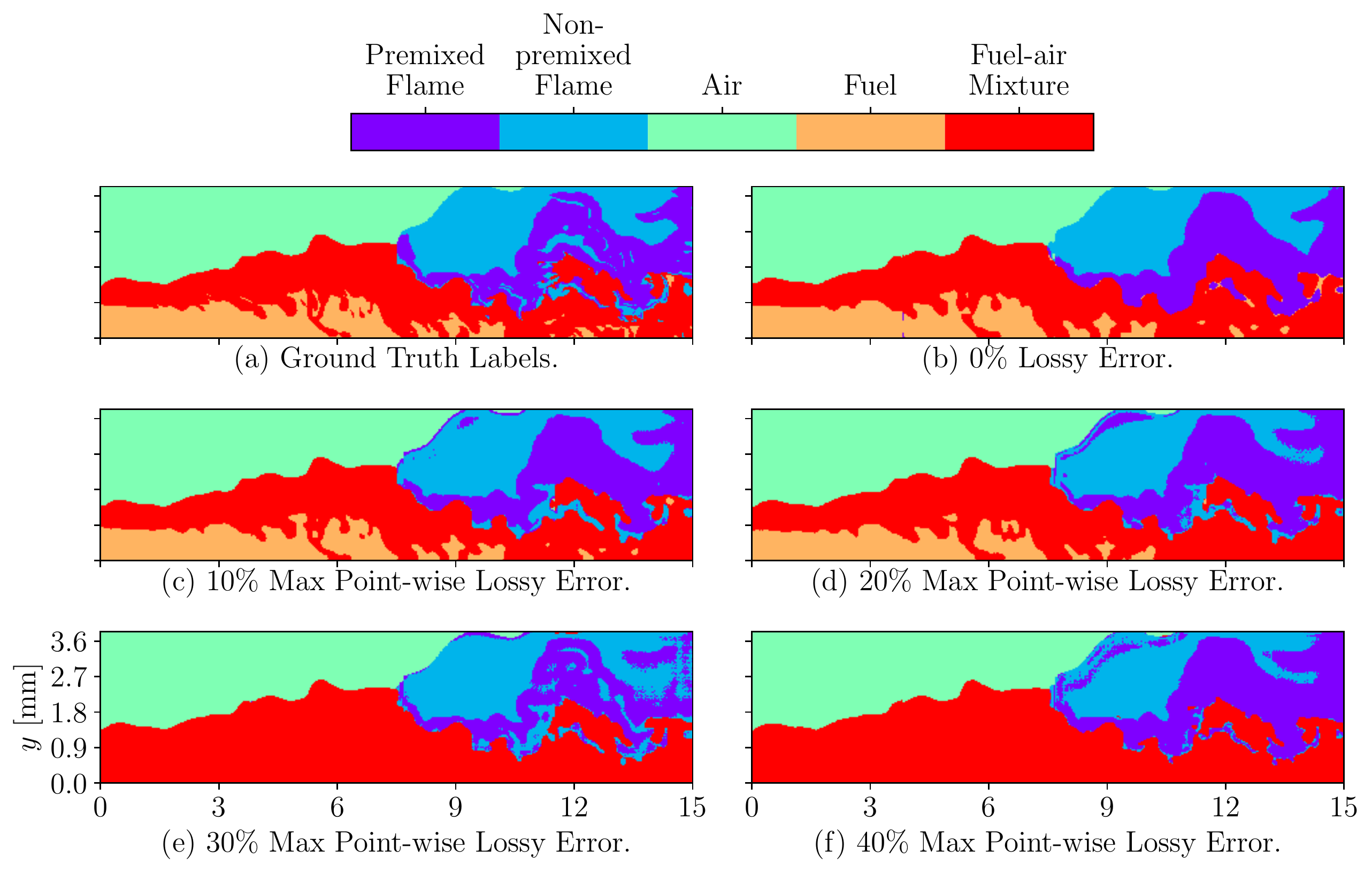}}
\caption{Visualization of ground truth and predictions from model trained on lossy features and lossy labels, and tested on clean features and labels.}
\label{fig:vis_noisy}
\end{center}
\vskip -0.4in
\end{figure*}

\subsubsection{Lossy Features and Lossy Labels}

\Cref{fig:noisy_acc} compares class-specific accuracy scores, along with the mean of these scores, for different levels of maximum point-wise lossy errors,when training with both lossy features and lossy labels. The mean accuracy scores are seen to be robust up to only 10\% max point-wise lossy error, which still corresponds to a 11-fold compression in the original data. After 20\% max lossy error, class accuracy for fuel drops promptly to 0, which can be explained by the highly distorted training labels for fuel shown in \Cref{fig:label_lossy}c. Remarkably, the deep learning model demonstrates reasonably robust behavior in the other classes, especially in the flame regions, to  max point-wise lossy errors up until 40\% .

\Cref{fig:vis_noisy} visualizes the predictions from the deep learning model trained on lossy features and lossy labels, and tested on clean features and labels. \Cref{fig:vis_noisy}b shows that the model predictions at 0\% lossy error are in reasonable agreement with the ground truth labels in \Cref{fig:vis_noisy}a. Evident misclassification of non-premixed flame is seen near the boundary with air in  \Cref{fig:vis_noisy}b and \Cref{fig:vis_noisy}c. This is likely caused by the presence of noisy labels between the premixed and non-premixed flame labels seen in \Cref{fig:label_lossy}. Nevertheless,  coherent classification is still observed in the flame regions,  despite the noisy training labels up to 40\% max point-wise lossy error in  \Cref{fig:label_lossy}c, as previously discussed with the class accuracy scores in \Cref{fig:noisy_acc}.

\section{Conclusions}
\label{sec:conclusion}
In this paper, we provide the requirements for establishing a realistic big data framework for scientific ML. These requirements are (i) community involvement, (ii) public data repositories, and (iii) lossy compression algorithms. We provide a proof-of-concept for this framework, which we name BLASTNet~\cite{BLASTWeb}, at \url{https://blastnet.github.io/}. 

To demonstrate that lossy data is useful for training ML algorithms, we compress data from a petascale simulation of a turbulent reacting flow configuration, and employ the compressed data to train a 3-D CNN in a semantic segmentation problem. Two scenarios are investigated: 
(i) where lossy features are shared into the repository with clean labels, and (ii) where lossy labels are generated from raw lossy data obtained from the repository. In the case of only lossy features, the CNN is robust up until 20\% max point-wise lossy error, corresponding to a compression of 13-fold. The CNN is robust up until 10\% max point-wise lossy error, corresponding to a 11-fold compression ratio. These results indicate that accurate predictions can still be made by deep learning algorithms even when training with lossy data, and that lossy compression can be utilized to mitigate storage constraints in open-source data repositories. 

We intend to extend the analysis presented here to consider more complex regression problems through future studies.  Nevertheless, the results from this work show that with community involvement, public data repositories, and lossy compression algorithms, the challenging task of creating and storing big data for scientific ML can be more bearable.

\section*{Acknowledgments}
The authors acknowledge funding support from the Department of Energy (DoE) Office of Basic Energy Sciences under award DE-SC002222. We are also grateful for financial support and computing resources from the DoE National Nuclear Security Administration, under award No. DE-NA0003968.
 The work at Sandia National Laboratories was supported by the DoE, Office of Basic Energy Sciences, Division of Chemical Sciences, Geosciences, and Biosciences. Sandia National Laboratories is a multimission laboratory managed and operated by National Technology and Engineering Solutions of Sandia, LLC., a wholly owned subsidiary of Honeywell International, Inc., for DoE National Nuclear Security Administration under contract DE-NA0003525.

\section*{Supplementary Material}

The code and models employed in this study can be found in \url{https://github.com/IhmeGroup/lossy_ml}. Lossy and clean data used in this study can be found in \url{https://www.kaggle.com/datasets/waitongchung/chung-et-al-icmlw-ai4science}. In addition, the links to proof-of-concept website proposed in \Cref{fig:proposal} is found in \url{https://blastnet.github.io/}, which also provides standards for contributing data and tutorials on reading and accessing shared data.


\bibliography{2_example_paper}
\bibliographystyle{icml2022}

%%%%%%%%%%%%%%%%%%%%%%%%%%%%%%%%%%%%%%%%%%%%%%%%%%%%%%%%%%%%%%%%%%%%%%%%%%%%%%%
%%%%%%%%%%%%%%%%%%%%%%%%%%%%%%%%%%%%%%%%%%%%%%%%%%%%%%%%%%%%%%%%%%%%%%%%%%%%%%%
% APPENDIX
%%%%%%%%%%%%%%%%%%%%%%%%%%%%%%%%%%%%%%%%%%%%%%%%%%%%%%%%%%%%%%%%%%%%%%%%%%%%%%%
%%%%%%%%%%%%%%%%%%%%%%%%%%%%%%%%%%%%%%%%%%%%%%%%%%%%%%%%%%%%%%%%%%%%%%%%%%%%%%%
\newpage
\appendix
\onecolumn
\section{Training and Validation}
\label{app:trainval}

\begin{figure}[htb!]
        \centering
            \begin{subfigure}{0.45\columnwidth}
        \centering            \includegraphics[width=\columnwidth]{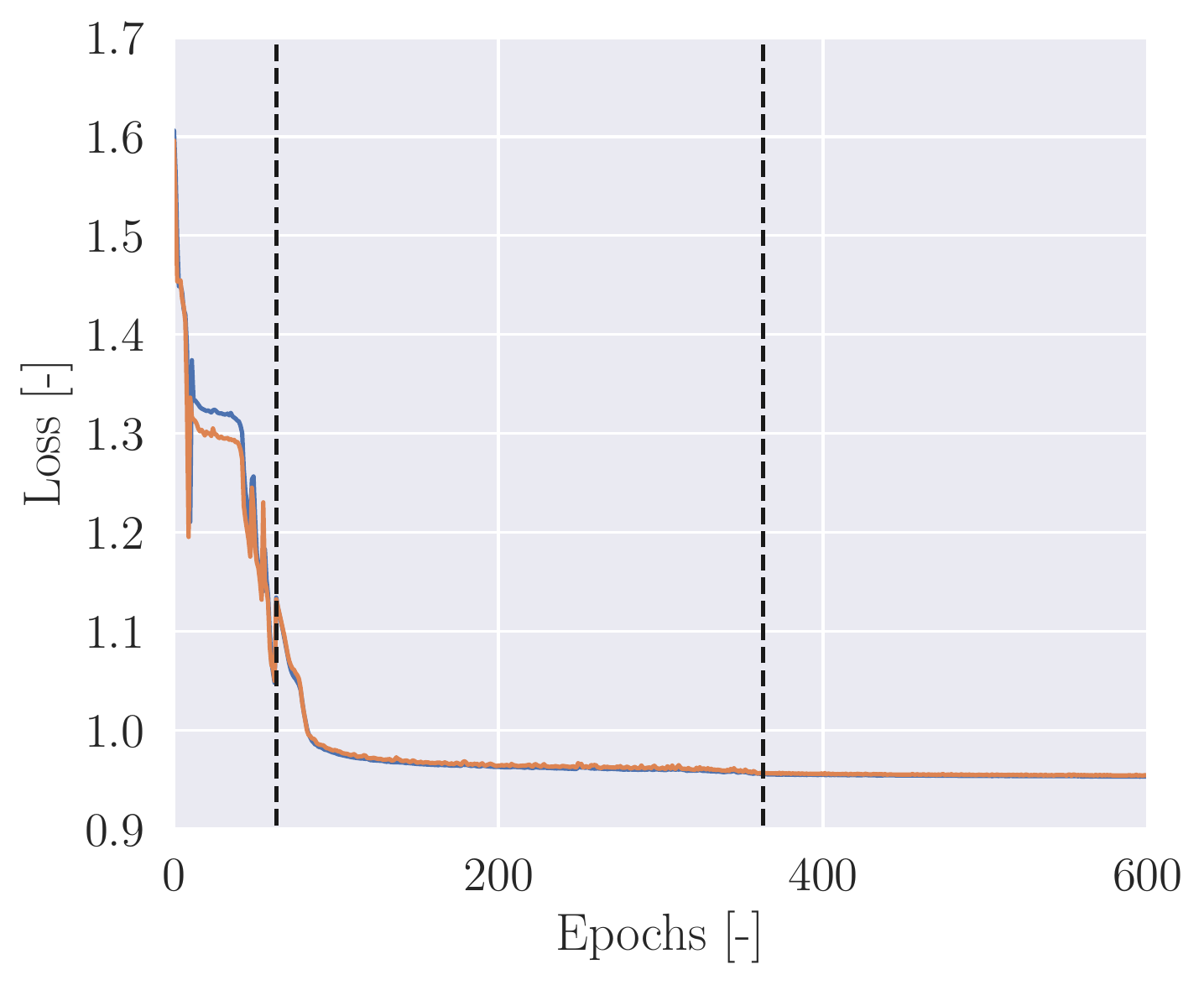}
        \caption{20\% max  point-wise lossy error in only features.\label{fig:loss_clean20}}
    \end{subfigure}
    \begin{subfigure}{0.45\columnwidth}
        \centering            \includegraphics[width=\columnwidth]{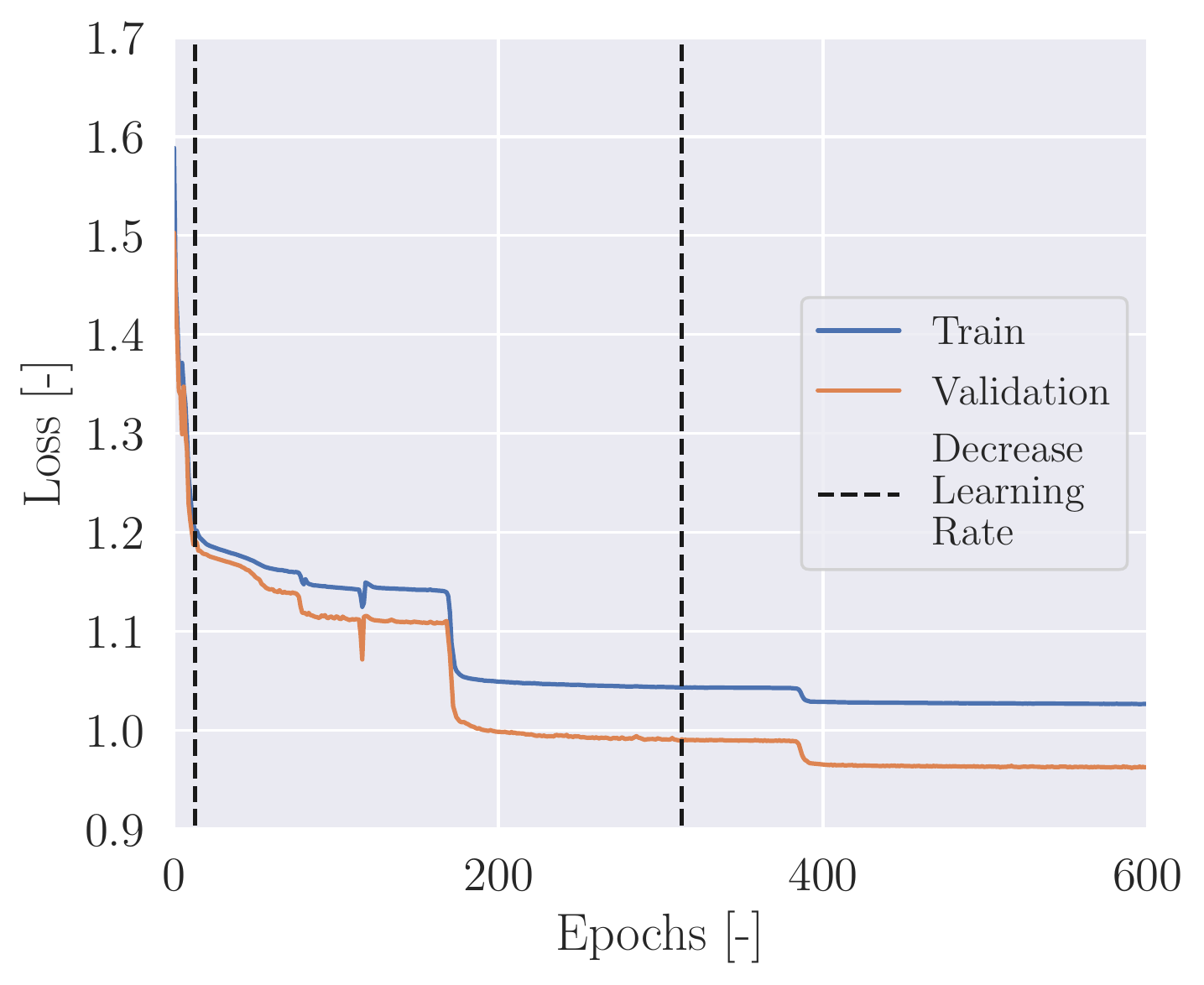}
        \caption{10\% max point-wise lossy error in both features and labels.\label{fig:loss_noise}}
    \end{subfigure}

    \caption{Loss during training.\label{fig:loss}}
\end{figure}

Training is performed with the Adam~\cite{Kingma2014} optimizer, with a batch size of 24 and raw learning rates of 1E-4, 1E-5, and 1E-6 for 100, 300, and 300 epochs, respectively, with  early-stopping employed when necessary. Prior to training, the raw learning rates are multiplied by the square root of the batch size. Note that in \Cref{fig:loss_noise}, the converged validation loss can be lower than the training loss, leading to higher validation accuracy than training accuracy. This is caused by the absence of lossy errors in the validation set, as described in \Cref{sec:results}.
Training this model on four Tesla V100 GPUs requires a total of approximately 4 hours of wall-clock-time for each case. 

%%%%%%%%%%%%%%%%%%%%%%%%%%%%%%%%%%%%%%%%%%%%%%%%%%%%%%%%%%%%%%%%%%%%%%%%%%%%%%%
%%%%%%%%%%%%%%%%%%%%%%%%%%%%%%%%%%%%%%%%%%%%%%%%%%%%%%%%%%%%%%%%%%%%%%%%%%%%%%%

\end{document}